\documentclass[a4paper, 10pt, conference]{ieeeconf}      % Use this line for a4 paper

\IEEEoverridecommandlockouts                              % This command is only needed if 
\usepackage{graphics} % for pdf, bitmapped graphics files
\usepackage{epsfig} % for postscript graphics files
\usepackage{mathptmx} % assumes new font selection scheme installed
\usepackage{times} % assumes new font selection scheme installed
\usepackage{amsmath} % assumes amsmath package installed
\usepackage{amssymb}  % assumes amsmath package installed

\usepackage{subcaption}
\usepackage{url}
\title{\LARGE \bf
Development of a 3 in Sewer Pipe Inspection Robot with an Articulated Differential Mechanism using X-shaped Linkages
}

\author{Shoya Umemura, Ryota Taniguchi and Atsushi Kakogawa% <-this % stops a space
\thanks{*This work was supported by KSAC (Kansai Startup Academia Coalition) GAP Fund, Japan Science and Technology Agency (JST)}% <-this % stops a space
\thanks{S. Umemura, R. Taniguchi and A. Kakogawa are with the Department of Robotics, Faculty of Science and Engineering, Ritsumeikan University, 1-1-1 Noji-higashi, Kusatsu, Shiga, 525-8577, JAPAN
        {\tt\small rr0146iv@ed.ritsumei.ac.jp, rr0145hv@ed.ritsumei.ac.jp, kakogawa@fc.ritsumei.ac.jp}}%
}

\begin{document}

\maketitle
\thispagestyle{empty}
\pagestyle{empty}

\begin{abstract}
This paper proposes, an improved version of the 3 in sewer pipe inspection robot equipped with an emergency evacuation mechanism. The low traction force and poor step-over capability, which were challenges of the first version, have been improved by simply connecting the propulsion units. The coupled propulsion units feature a differential mechanism capable of posture changes via a single wire, enabling adaptation to pipe diameter variations. To traverse obstacles like pipe joints, a control method was devised that detects obstacle contact through current load on the drive wheel motors and slackens the wire. This method was verified through simulated pipe experiments. Load comparisons were made using current waveforms applied to the drive wheels. Our proposed control method significantly improved the step-over capability of the new articulated robots.
\end{abstract}

%%%%%%%%%%%%%%%%%%%%%%%%%%%%%%%%%%%%%%%%%%%%%%%%%%%%%%%%%%%%%%%%%%%%
\section{INTRODUCTION}
%研究背景
In recent years, accidents caused by aging sewer pipelines have become a social issue in Japan. In 2024, the total length of sewer pipelines over 50 years old in Japan is approximately 39,200 km \cite{MLIT}, and this number is increasing, necessitating urgent pipeline inspections.
%近年，下水道配管の老朽化による事故の発生が社会問題となっている．2024年時点で敷設から50年が経過した下水配管の総延長は39200kmに上り，増加傾向にある．このため，早急に配管点検を実施する必要がある．
However, it is difficult to inspect narrow inner-diameter pipelines with 100 mm or less.
While push-in cameras \cite{in_pipe_camera} are a basic way to inspect these sewer pipelines, their high cable bending stiffness makes navigating multiple bends challenging.
%しかし，内径が100mm以下となる様な配管では，現状検査を行う事が困難である．こうした配管には従来押し込み式カメラが用いられているが，ケーブルの曲げ剛性が非常に高いため，複数の曲管を含む配管経路の通過が困難である．
Consequently, there is a strong demand for self-propelled inspection robots capable of actively traversing and inspecting inside narrow inner diameter pipelines. 
%以上から，小口径配管を能動的に走行し点検が可能な小口径配管検査ロボットに注目が集まっている（渇望されている？）．

Existing examples of narrow inner diameter pipe inspection robots include a peristaltic motion robot using pneumatic actuators \cite{Nakamura1, Nakamura2} and a 1-in pipe inspection robot \cite{TOSHIBA} that travels by pressing the wheels against the inner wall of the pipe using a planetary gear structure. 
%既に報告されている小口径配管検査ロボットの例としては，空気圧アクチュエータを用いた蠕動運動型ロボットや，遊星歯車機構により車輪を配管内壁に押し付けて走行する１インチ配管検査ロボットが挙げられる．
However, the peristaltic motion robot moves at low speeds of approximately 0.01 m/s and requires a large, heavy compressor for operation. In the case of sewer inspections, time constraints exist due to road closures, and there are sites where large, heavy equipment cannot be transported. Therefore, this method cannot be considered practical in all cases.
1-in inspection robot has a high risk due to its complex design using planetary gear structures. It constantly presses its wheels inner wall of the pipe during operation. Consequently, if an unexpected urgent accident occurs during operation, evacuation of the robot and withdrawal become difficult.
%しかし，前者は動作スピードが遅く，動作の為に大型で大重量なコンプレッサを必要とするため，運用性に欠ける．また，後者についても遊星歯車機構という複雑な機構をしているため故障リスクが高く，常に配管内壁に車輪を押し付けて走行するため回収性に欠けるという課題がある．

Previously, we have developed a wire-driven 3 in sewer pipe inspection robot with a wire-driven parallel elastic actuator for emergency evacuation, named Xbot-1 \cite{Xbot-1}. 
It combined high speed with a capability of withdrawal. However, having only one propulsion unit resulted in insufficient traction over extended distances, limiting its inspection range. Furthermore, the propulsion units tended to get stuck when passing through steps such as pipe joint.
%以上から，筆者らはワイヤ駆動機構による緊急脱出機構を有する3インチ配管内検査ロボットXbot-1を開発した．高い動作スピードと回収性を両立するものであったが，推進ユニットが１基のみのため検査距離が延びるにつれ牽引力が不足し検査可能距離に制限があった．更に，配管継手などの段差を走行する際，推進ユニットがスタックしてしまうという課題があった．

\begin{figure}[t]
    \centering
    \includegraphics[width=0.9\linewidth]{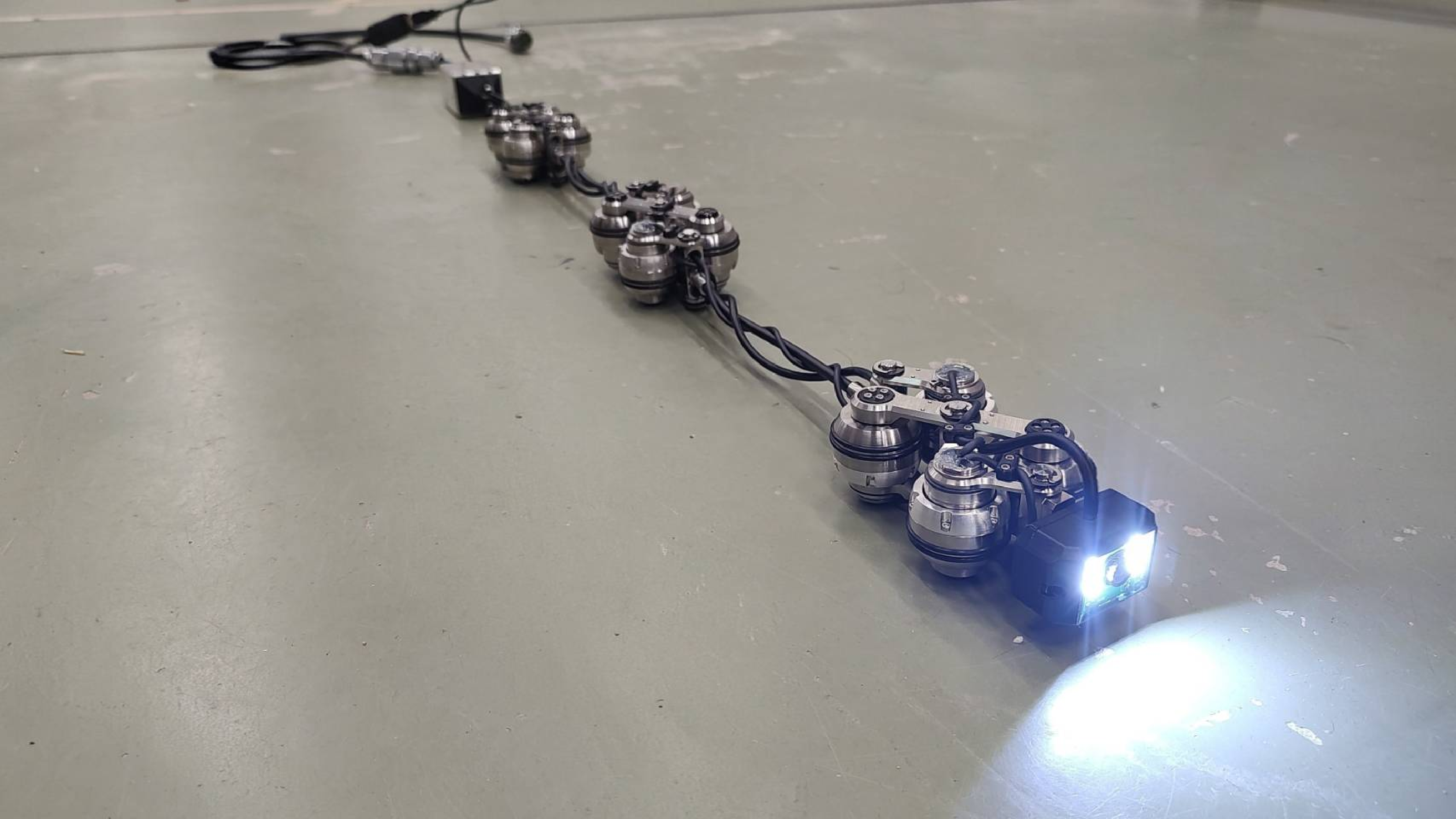}
    \caption{Overview of 3 in sewer pipe inspection robot ``Xbot-2''}
    \label{Xbot-2}
    \vspace{-5mm}
\end{figure}

This paper proposes the second version of Xbot-1, named Xbot-2 (Fig.~\ref{Xbot-2}), a 3-in pipe inspection robot with an articulated differential mechanism of X-shaped links that improves the traction force problem by connecting propulsion units. Furthermore, a control method for passing through steps such as pipe joints was proposed, and their effectiveness was also verified by experiments with an actual robot and pipes.
%本研究では，Xbot-1の課題であった牽引力を推進ユニットの連結により改善した連結差動機構を有する3インチ配管内検査ロボットXbot-2を提案する．また，複数の推進ユニットを駆使して継手段差を走破可能とする制御を考案し，模擬配管における実験によりその効果を検証した.

%%%%%%%%%%%%%%%%%%%%%%%%%%%%%%%%%%%%%%%%%%%%%%%%%%%%%%%%%%%%%%%%%%%%
\section{MECHANICAL STRUCTURE}
%ロボットの機構
%===================================================================
\subsection{Robot configuration and motion principle}
\begin{figure}[tbp]
    \centering
    \includegraphics[width=\linewidth]{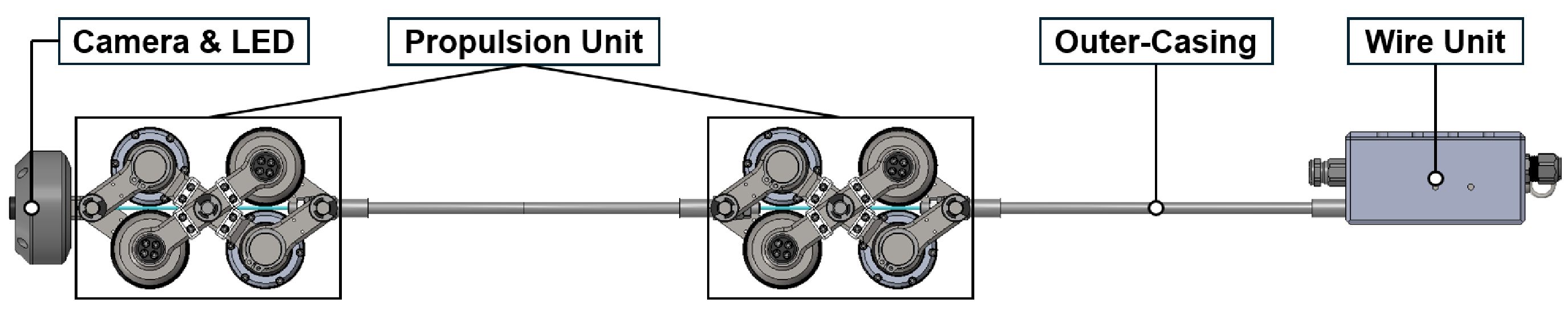}
    \caption{Overall configuration of the robot ``Xbot-2''}
    \label{Xbot_components}
\end{figure}

\begin{figure}[tbp]
  \begin{minipage}{0.55\columnwidth}
    \centering
    \includegraphics[width=\linewidth]{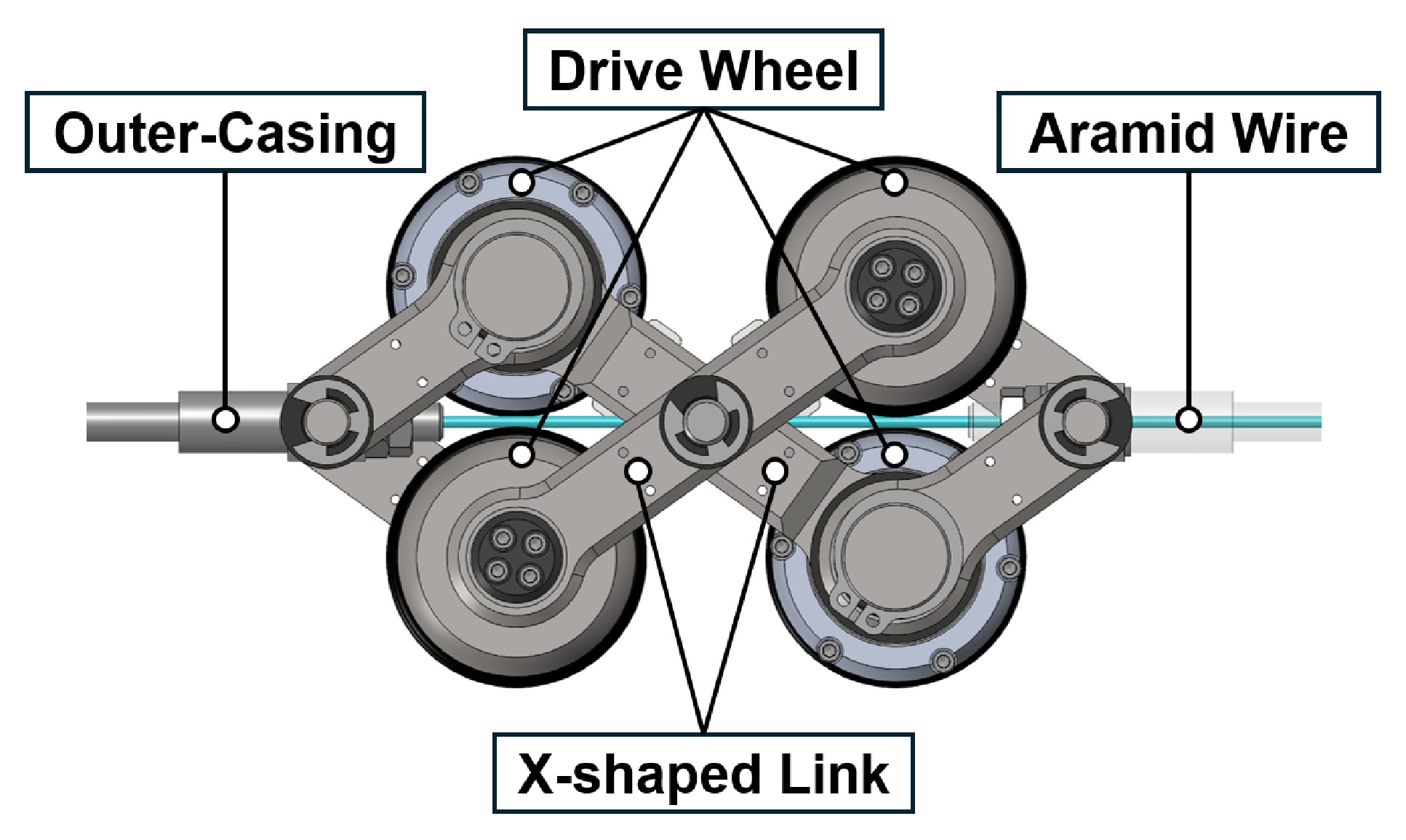}
    \subcaption{Propulsion unit}
  \end{minipage}
  \begin{minipage}{0.4\columnwidth}
    \centering
    \includegraphics[width=\linewidth]{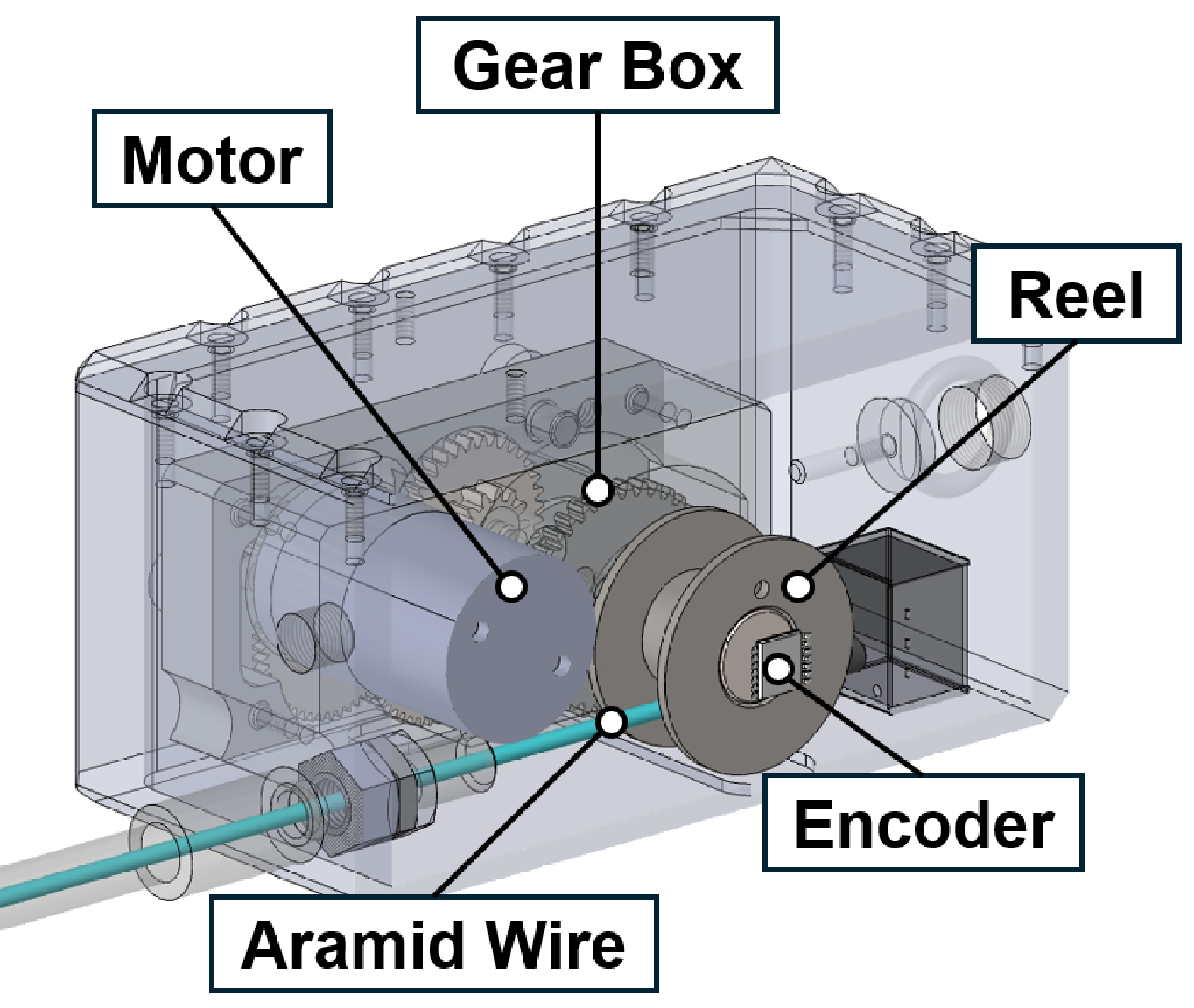}
    \subcaption{Wire-driven unit}
  \end{minipage}
  \caption{Mechanisms of two main drive unit}
  \label{Xbot_mechanism}
\end{figure}

%機体構成と動作原理
Overall configuration of Xbot-2 is shown in Fig.~\ref{Xbot_components}. It consists of a single front camera and LED, propulsion units, and a wire-driven unit connected by outer-casing. This robot can inspect the inside of the pipe by taking a video with the camera.
%機体構成をFig.に示す．本ロボットはカメラ・LED，推進ユニット，ワイヤユニットをアウタケーシングによって接続されて構成されている．カメラを用いた撮影により，配管内を検査可能である．
Figure \ref{Xbot_mechanism} shows detailed mechanisms of two main drive unit.
The propulsion unit is equipped with a total of four drive wheels with geared motors built into the end points of the X-shaped link mechanism. An aramid wire passes through the outer-casing, and when this is wound up by the wire unit, the length of the link mechanism's axis is shortened. As a result, the outer diameter of Xbot-2 expands in the pipe radius direction, then the wheels can be pressed.
Finally, the drive wheels generate a propulsion force, and the robot can move in the pipe.
%Fig.に機構の詳細と推進ユニットの動作原理を示す．推進ユニットはクロス状リンクとその末端に取り付けられた4つのギヤードモータを内蔵する駆動輪から構成される．アラミド繊維ワイヤが中心を貫通するように通っており，ワイヤユニットに内蔵されたギヤードモータ・リール機構によって巻き取られることでクロス状リンクの軸長が収縮する．これによりXbot-2の外径が配管半径方向に拡大し，駆動輪を配管内壁に押し付けられる．駆動輪が推進力を発生させることで配管内を移動可能となっている．

%===================================================================
\subsection{Articulated differential mechanism by connecting propulsion units}
\label{Differential_mechanism_by_connecting_propulsion_units}
\begin{figure}[tbp]
  \begin{minipage}{\columnwidth}
    \centering
    \includegraphics[width=0.5\linewidth]{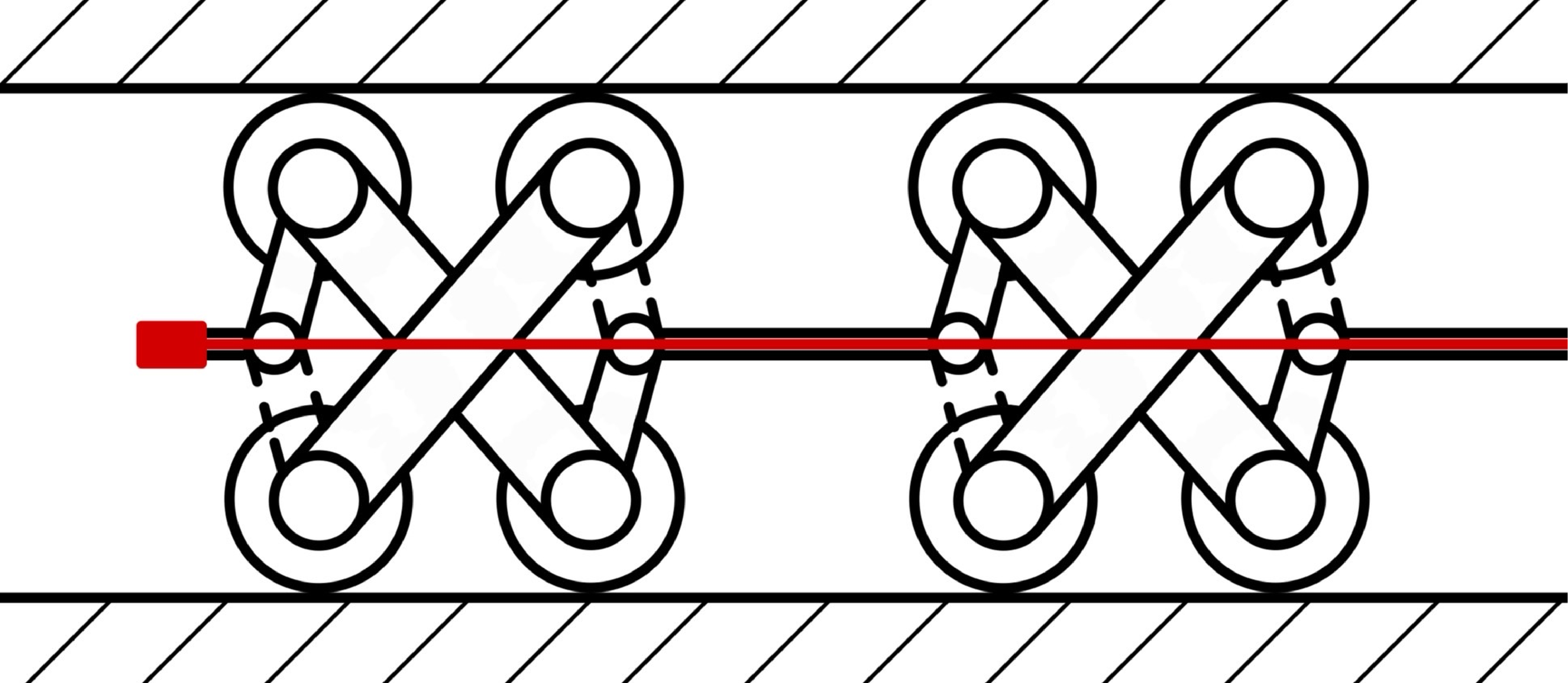}
    \subcaption{Pipe with constant inner diameter}
  \end{minipage}
  \\
  \begin{minipage}{\columnwidth}
    \centering
    \includegraphics[width=0.5\linewidth]{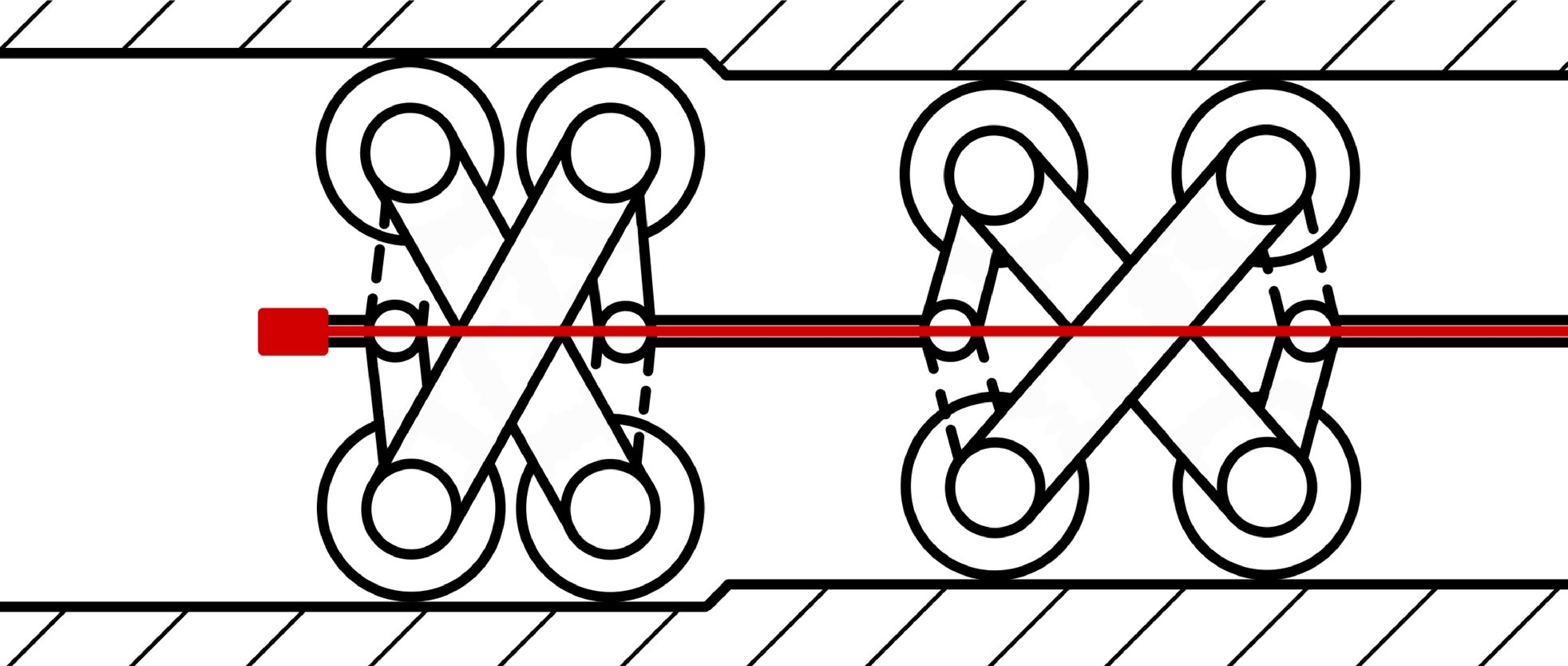}
    \subcaption{Expansion in inner diameter}
  \end{minipage}
  \\
  \begin{minipage}{\columnwidth}
    \centering
    \includegraphics[width=0.45\linewidth]{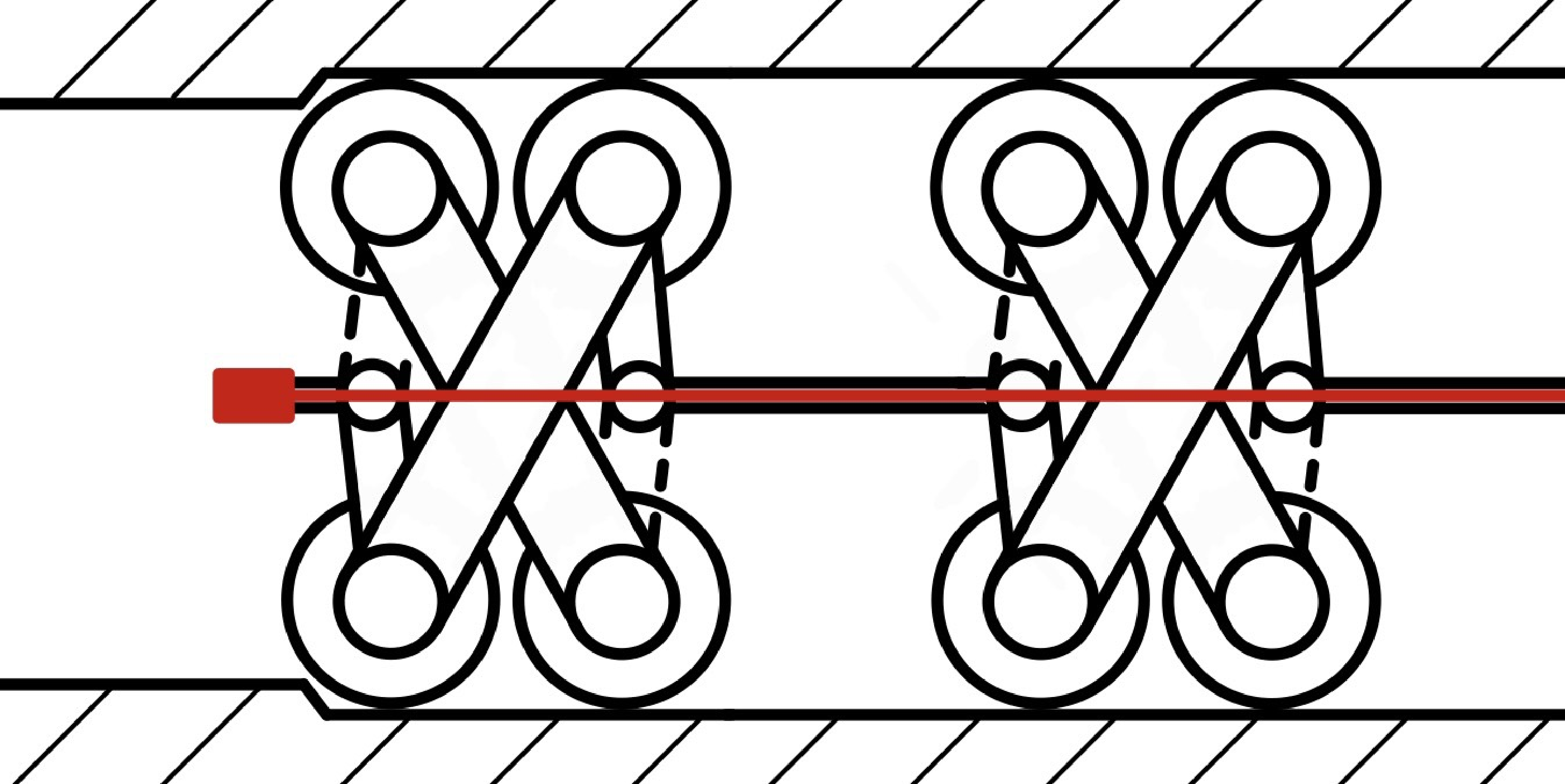}
    \subcaption{Shrinkage in inner diameter}
  \end{minipage}
  \caption{Motion principle of the differential mechanism in stepped pipes}
  \label{Differential_mechanism}
\end{figure}

%推進ユニットの連結による差動機構
Figure ~\ref{Differential_mechanism} shows a schematic diagram of the drive wheels pressed against the inner wall of the pipe. Figure ~\ref{Differential_mechanism} (a) represents a pipe with a constant inner diameter. In this case, the deformation amount of each X-shaped linkage is equal. On the other hand, as shown in Fig.~\ref{Differential_mechanism} (b), since the inner diameter of the pipe expands, the contact force from the inner wall of the pipe at this larger pipe decreases. As a result, only the X-shaped linkage of the front propulsion unit deforms by the wire tension.
The outputs (deformation amount of each X-shaped linkage) change depend on the external environment (how the inner space of the pipe is geometrically constrained at each propulsion unit).
This is precisely the same principle as a differential mechanism. By connecting three or more propulsion units, an articulated differential mechanism \cite{Hirose} can be formed.
%配管内壁へ駆動輪を押し付けた状態の模式図をFig.3に示す．Fig.3(a)の状況は配管内径が一様な状況である．この場合，各推進ユニットの位置において配管内径は等しく，各クロス状リンクの変形量は一致する．一方，Fig.3(b)に示す様に途中で内径が拡大する場合では，配管壁面からの接触力が解消されることでワイヤ張力の作用を受けて前方の推進ユニットのみクロス状リンクが変形する．これは，入力（巻き取り機構によるワイヤ変位量）に対し，各出力（クロス状リンクの変形量）が外部環境（配管内径の変化）によってそれぞれ変化する差動機構である．推進ユニットを3基以上連結することにより，連結差動機構を構成可能である．

On the other hand, as shown in Fig.~\ref{Differential_mechanism} (c), it has a risk of getting stuck at the step of the pipe joint section when the inner diameter changes from that in Fig.~\ref{Differential_mechanism} (b).
If the wire tension is loosened, and the X-shaped linkage can be shrunk by the contact force from the pipe joint step.
%一方で，図2.6に示す様に，図2.5(b)の内径変化時から再び間内径が細くなる場合，上記の機構は配管継手部段差でスタックするリスクが考えられる．そこで，ワイヤを弛緩させる操作を提案する．これにより，推進ユニットの駆動輪が段差からの接触力によって姿勢変化可能になると考えられる．

%===================================================================
\subsection{Traction force measurement}
\begin{figure}[tbp]
    \centering
    \includegraphics[width=\linewidth]{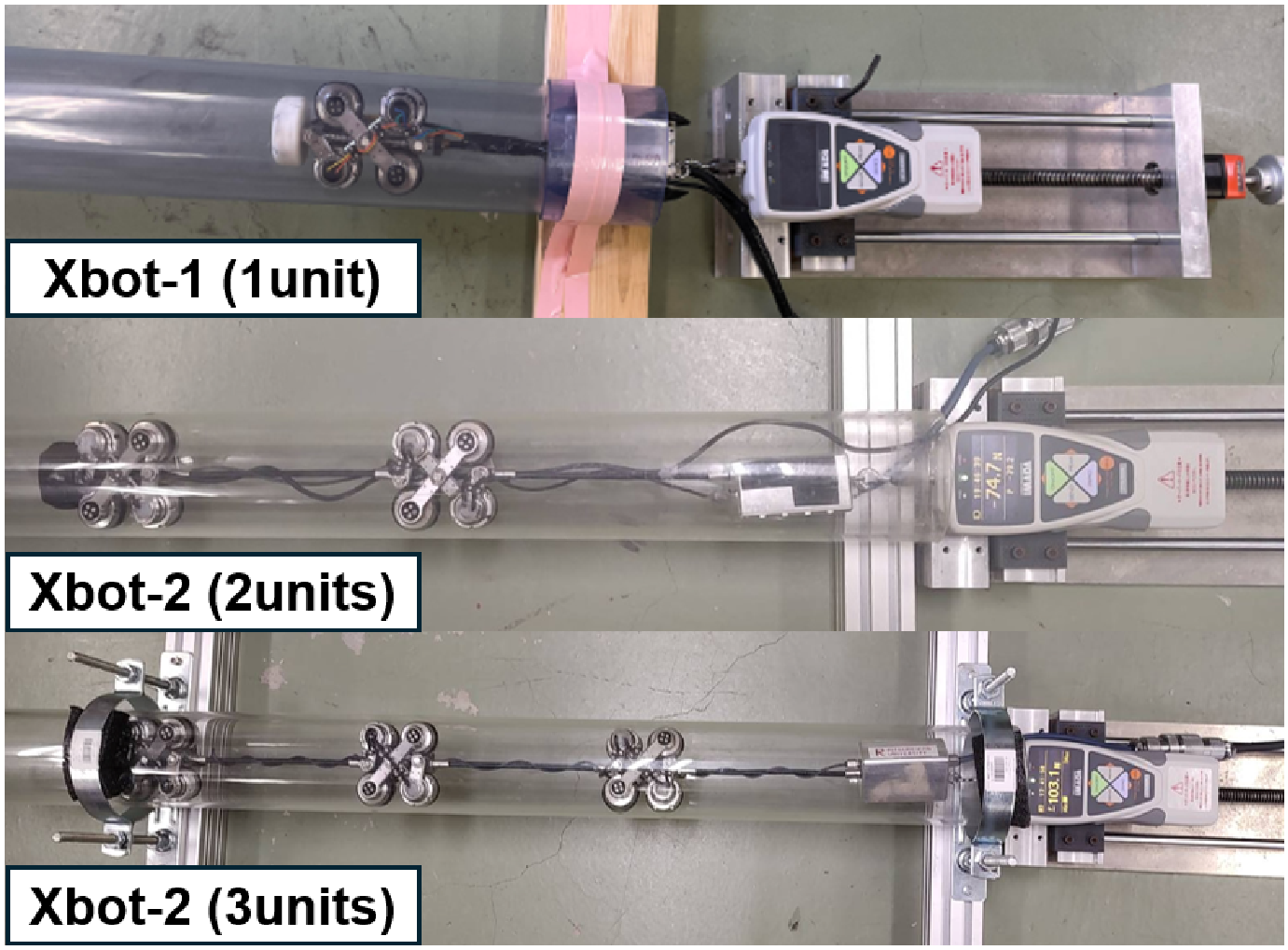}
    \caption{Traction force measurement with different propulsion unit number}
    \label{Traction_force_measurement}
\end{figure}

%牽引力測定
As shown in Fig.~\ref{Traction_force_measurement}, the traction force measurement was conducted with Xbot-1 and Xbot-2. The results are presented in Table~\ref{Comparison_of_traction_force}. The Xbot-2 has a bigger traction force compared to Xbot-1. In the case of two propulsion units, it has about 1.5 times the traction force; in the case of three propulsion units, it has about 2.0 times the traction force. Consequently, it was shown that the traction force is improved by connecting propulsion units.
%推進ユニットの連結による牽引力の変化を確認するため，Fig.に示す様にXbot-1，Xbot-2の牽引力測定を行った．得られた結果をTableに示す．Xbot-1と比較して，推進ユニットを２基連結した場合では1.5倍程度，3基連結した場合では2.0倍程度の牽引力を示した．これにより，Xbot-1の課題であった牽引力を改善できていることが確認された．

\begin{table}[tbp]
  \centering
  \caption{Measured traction force with different propulsion unit number}
  \label{Comparison_of_traction_force}
  \begin{tabular}{|c||c|c|c|}
    \hline
    Ver. & Xbot-1 & Xbot-2 (2 units) & Xbot-2 (3 units)\\
    \hline
    Traction force [N] & 50.6 & 75.0 & 103.1\\
    \hline
  \end{tabular}
\end{table}

%%%%%%%%%%%%%%%%%%%%%%%%%%%%%%%%%%%%%%%%%%%%%%%%%%%%%%%%%%%%%%%%%%%%
\section{AUTONOMOUS NEGOTIATION OF PIPE JOINTS BY STEP CONTACT DETECTION}
%段差接触検知による継手の自律走行

%===================================================================
\subsection{Preliminary test in straight pipes connected by a joint}
\begin{figure}[tbp]
    \centering
    \includegraphics[width=\linewidth]{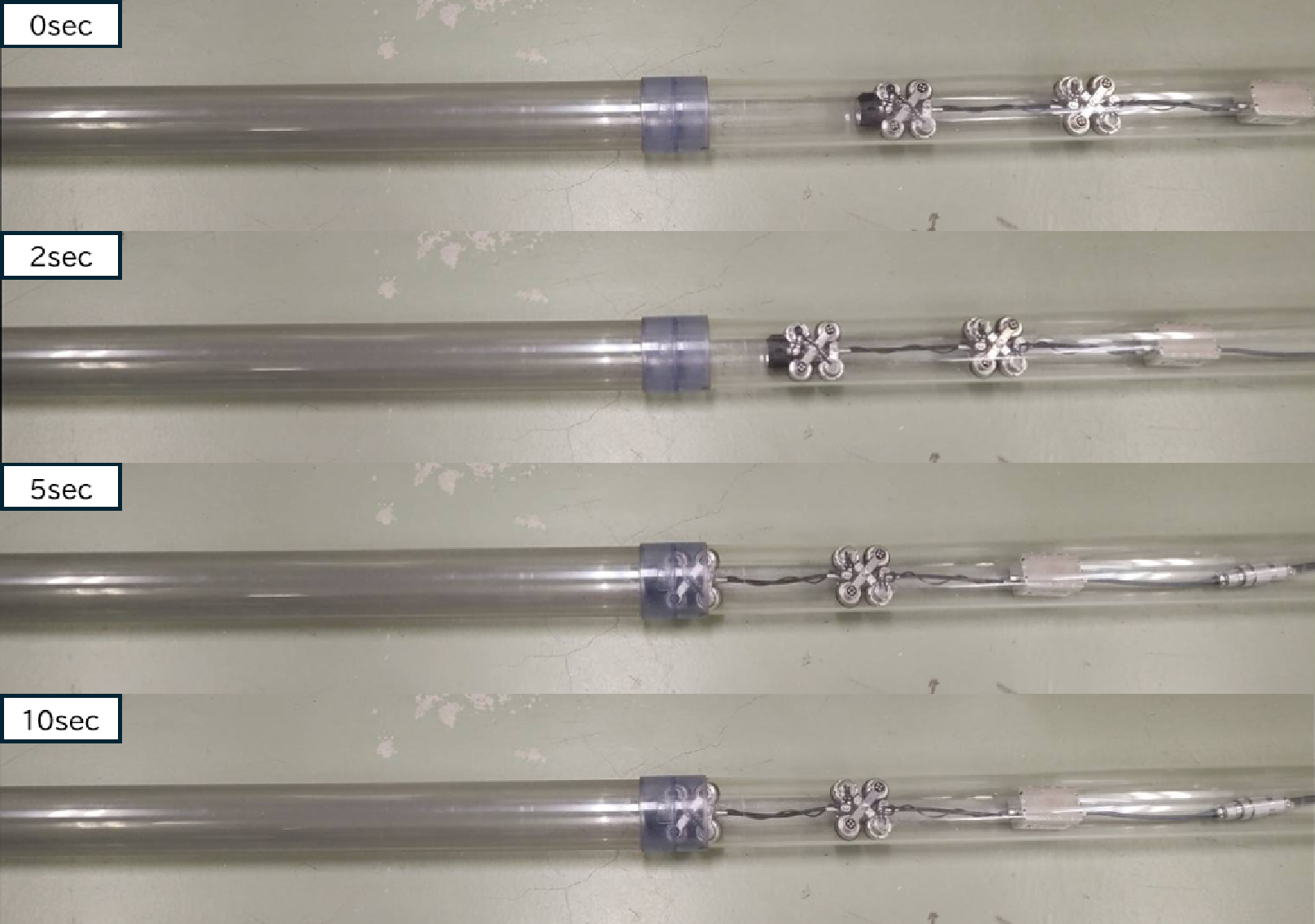}
    \caption{Transition of the actual motion with two propulsion units}
    \label{Joint_Traction_fail}
\end{figure}

\begin{figure}[tbp]
  \begin{minipage}{\columnwidth}
    \centering
    \includegraphics[width=\linewidth]{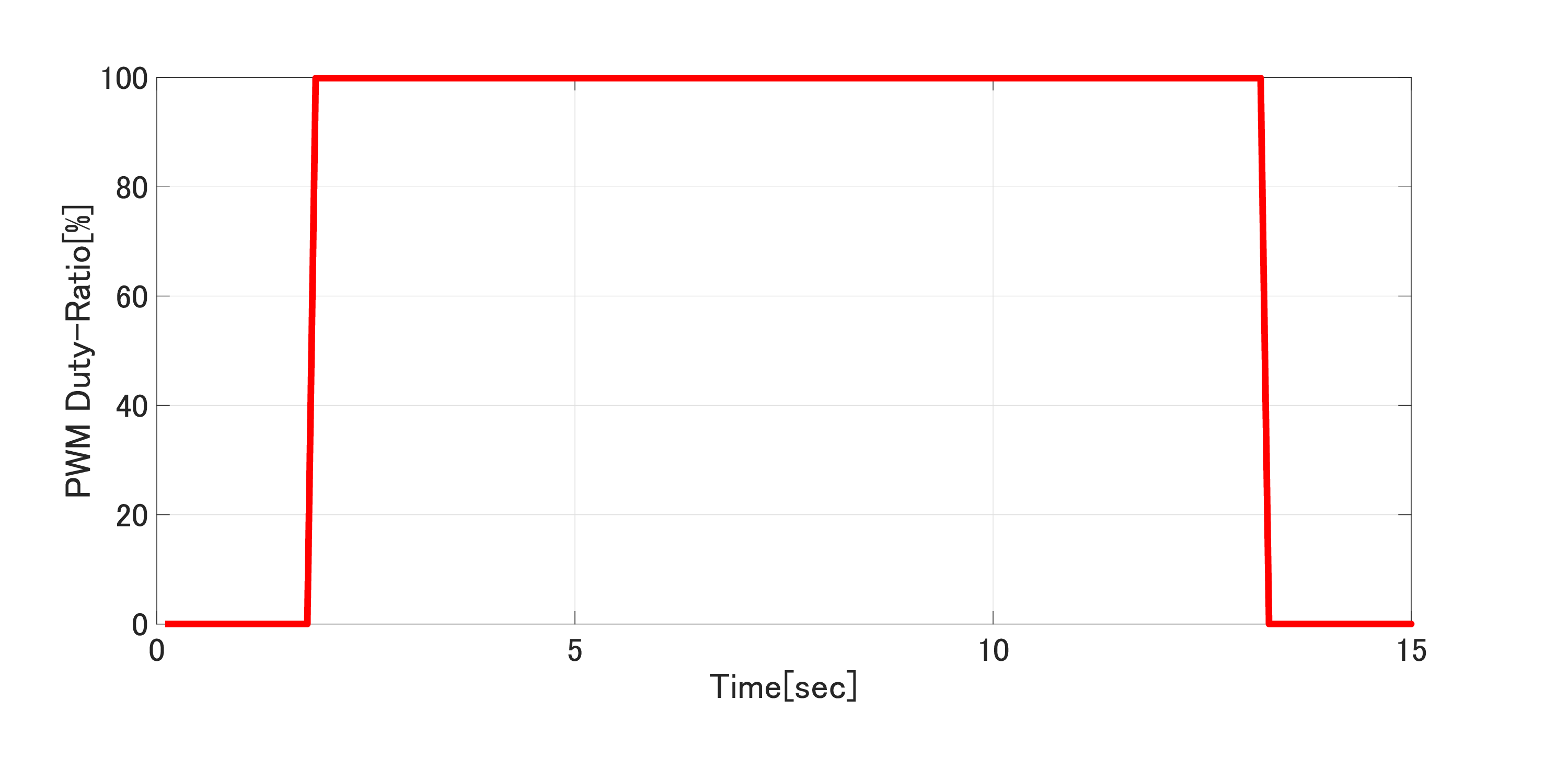}
    \vspace{-25pt}
    \subcaption{Duty ratio of the PWM input}
  \end{minipage}
  \\
  \begin{minipage}{\columnwidth}
    \centering
    \includegraphics[width=\linewidth]{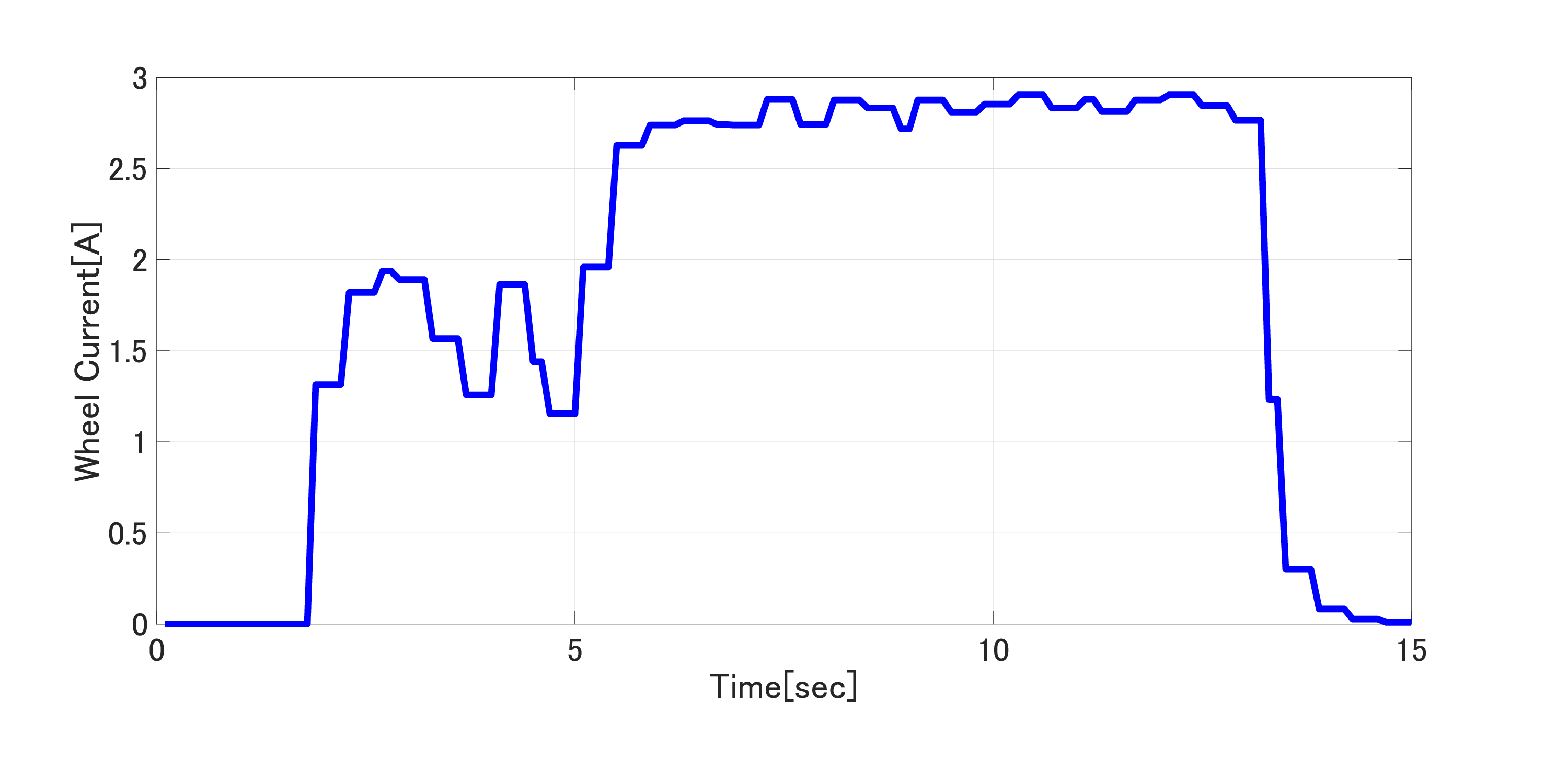}
    \vspace{-25pt}
    \subcaption{Drive wheel motor current}
  \end{minipage}
  \caption{Log data in motion with constant wire tension of the reel}
  \label{log_data_fail}
  %\vspace{-5mm}
\end{figure}

%模擬配管での走行検証
\label{Driving_test_in_an_imitation_pipe}
As discussed in the ~\ref{Differential_mechanism_by_connecting_propulsion_units}, the poor driving performance of our robot at the pipe joint step was examined by using two straight pipes connected by a single joint and Xbot-2 with two units.
Figure ~\ref{Joint_Traction_fail} shows the transition of the actual motion.
Figure~\ref{log_data_fail} plots the input PWM duty ratio for the drive wheels and their measured motor current.
%前節で想定した配管継手段差における走破性の低下について，模擬配管コースを用いて検証した．模擬配管コースは，アクリル製3インチ直管2本を配管継手で接続したものを使用した．走行時の様子を図に，スピード指令値，能動車輪モータを図に示す．

When the front propulsion unit contacted the pipe joint step, it immediately stopped. Although the PWM command value of 100 $\%$ was applied to the drive wheels continuously, the motor stalled, and rotation stopped.
Regarding the drive wheel motor current, it fluctuated between approximately 1.2 to 2.0 A until the contact with the pipe joint step (5 seconds).
However, after the contact, it remained at approximately 2.8 to 2.9 A.
%Xbot-2を模擬配管コースで走行させたところ，前方の推進ユニットが配管継手部の段差に接触し停止した．駆動輪に対しスピード指令値100%を与え続けたが，モータがストールし回転停止した．また，駆動輪モータ電流値について，継手段差に接触以前は1.2～2.0A程度で推移したが，接触後(5sec)以降は2.8～2.9A程度と高止まりした．

%===================================================================
\subsection{Algorithm to overcome pipe joint steps}
%配管継手における段差走破用制御プログラムの構築
An algorithm for pipe joint steps is validated to demonstrate the effectiveness of the loosening the wire tension when overcoming pipe joint steps.
%追記：継手段差を走破する際のワイヤの弛緩動作の有効性を示すため，継手自律走行用プログラムを構築し，検証した．
%-------------------------------------------------------------------
\subsubsection{Method to detect pipe joint steps}
%配管継手段差の検知手法
Autonomous navigation at joint sections requires detecting contact with steps.
When a step is encountered while the linkage is sufficiently pressed against the pipe, the rotation of the active wheel motor stops.
One might consider detecting this using encoders or an IMU.
However, these methods tend to increase the size of the propulsion unit and control board.
Therefore, this research implemented control that detects step contact without additional sensors by setting a threshold for the drive wheel motor current, thereby reducing wire tension.
%プログラムによる継手の自律走行には，段差との接触検知が必須となる．配管に対し十分にリンクを押し付けた状態で段差に接触すると，能動車輪モータがストールし回転が停止するため，エンコーダやIMUを用いたオドメトリが望ましい．しかし，これらの手法は推進ユニットや制御基板の大型化を必要とする．そこで，駆動輪モータ電流に閾値を設定することにより，段差との接触を検知してワイヤ張力を弱める制御を実装した．前節に示したように駆動輪が段差に接触してモータ負荷が高まり電流が上昇する現象を利用している．

%-------------------------------------------------------------------
\subsubsection{Control method to pass through pipe joint steps}
\begin{figure}[tbp]
    \centering
    \includegraphics[width=0.55\linewidth]{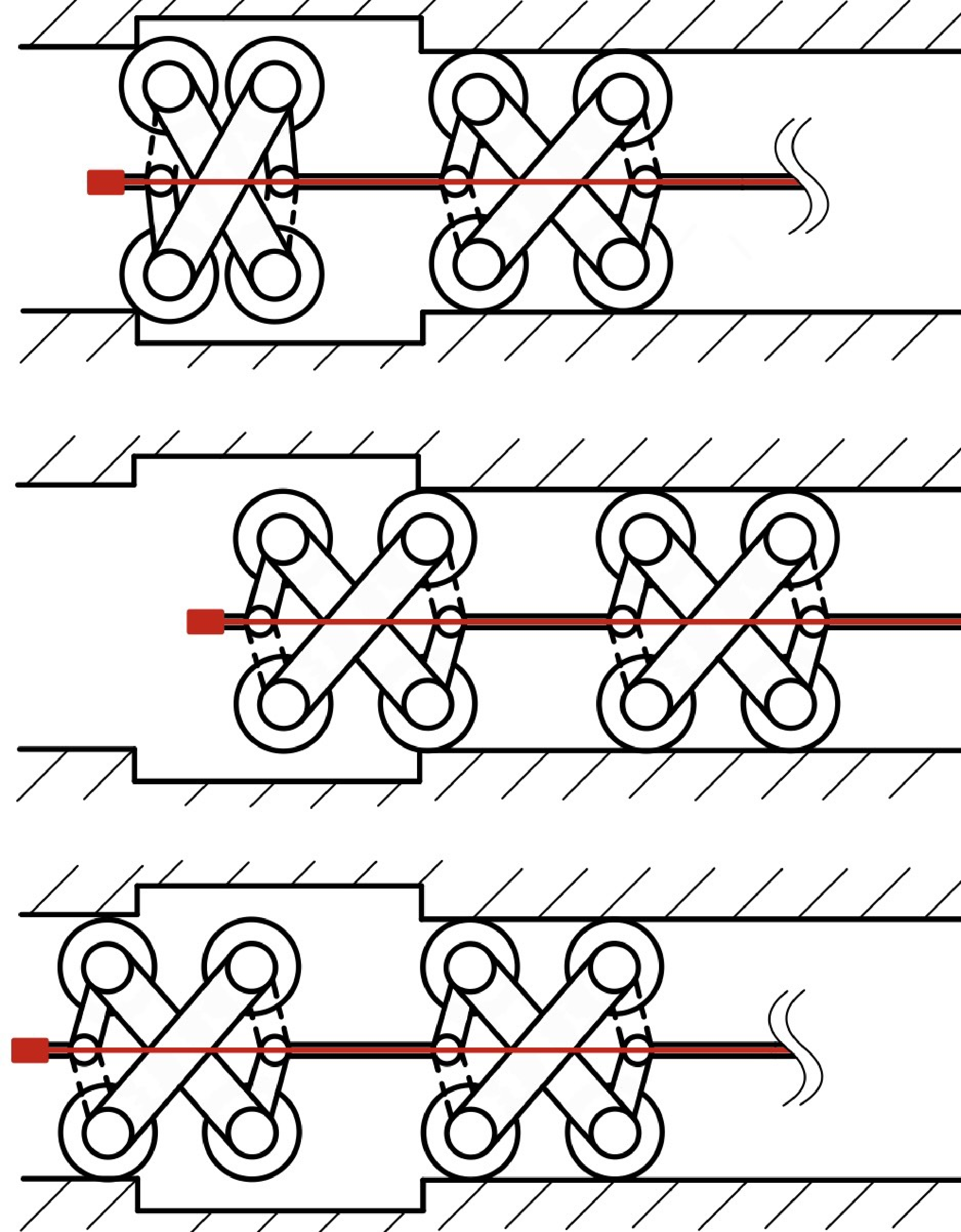}
    \caption{Schematic images of the control method to pass through pipe joint steps}
    \label{Control_method_image}
\end{figure}

%継手段差走破のための制御手法
Figure~\ref{Control_method_image} shows the schematic images of the control method to pass through pipe joint steps.
Its sequence is listed below:
%模擬配管での走行実験で得られた結果を元に，段差走破の為により効果的なロボット操作手法を考案した．図に走行時の動作イメージを示す．
\begin{enumerate}
  \item Contact detection by the wheel motor current threshold of 2.5 A
  %駆動輪電流の閾値（2.5A）による接触検知
  \item Loosening the wire tension by 150 deg reversal rotation of winding-up reel
  %角度制御により巻き取りリール角度を150°逆転させ，ワイヤを弛緩させる
  \item Backward movement for 0.5 sec
  %0.5sec後退する
  \item Shrinkage of the X-shaped linkage in the pipe radius direction by the contact force from the pipe joint step
  %配管継手段差から接触力を受け，リンクが配管半径方向に収縮する
  \item Forward movement and passing through the pipe joint
  %リンクの幅を保ったまま前方の推進ユニットが配管継手を通過する
\end{enumerate}

%===================================================================
\subsection{Autonomous driving using the proposed control method to pass through pipe joint step}
\begin{figure}[tbp]
  \centering
  \includegraphics[width=\linewidth]{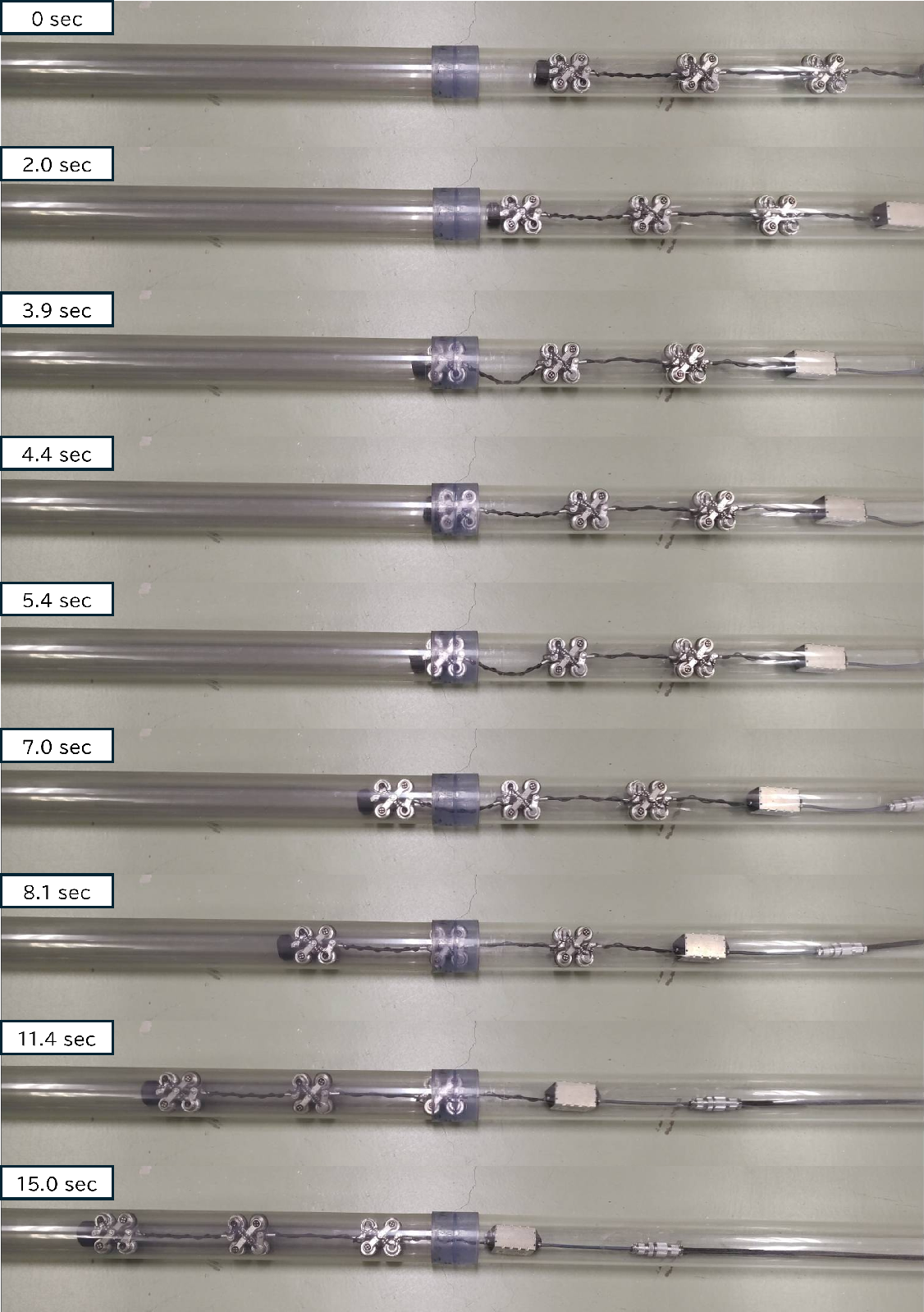}
  \caption{The transition of the actual motion with three propulsion
units and the proposed algorithm of the wire-driven mechanism}
  \label{fig:3PropulsionUnit_Exp3_picture}
  %\vspace{-5mm}
\end{figure}

\begin{figure}[tbp]
    \vspace{-5mm}
  \begin{minipage}{\columnwidth}
    \centering
    \includegraphics[width=\linewidth]{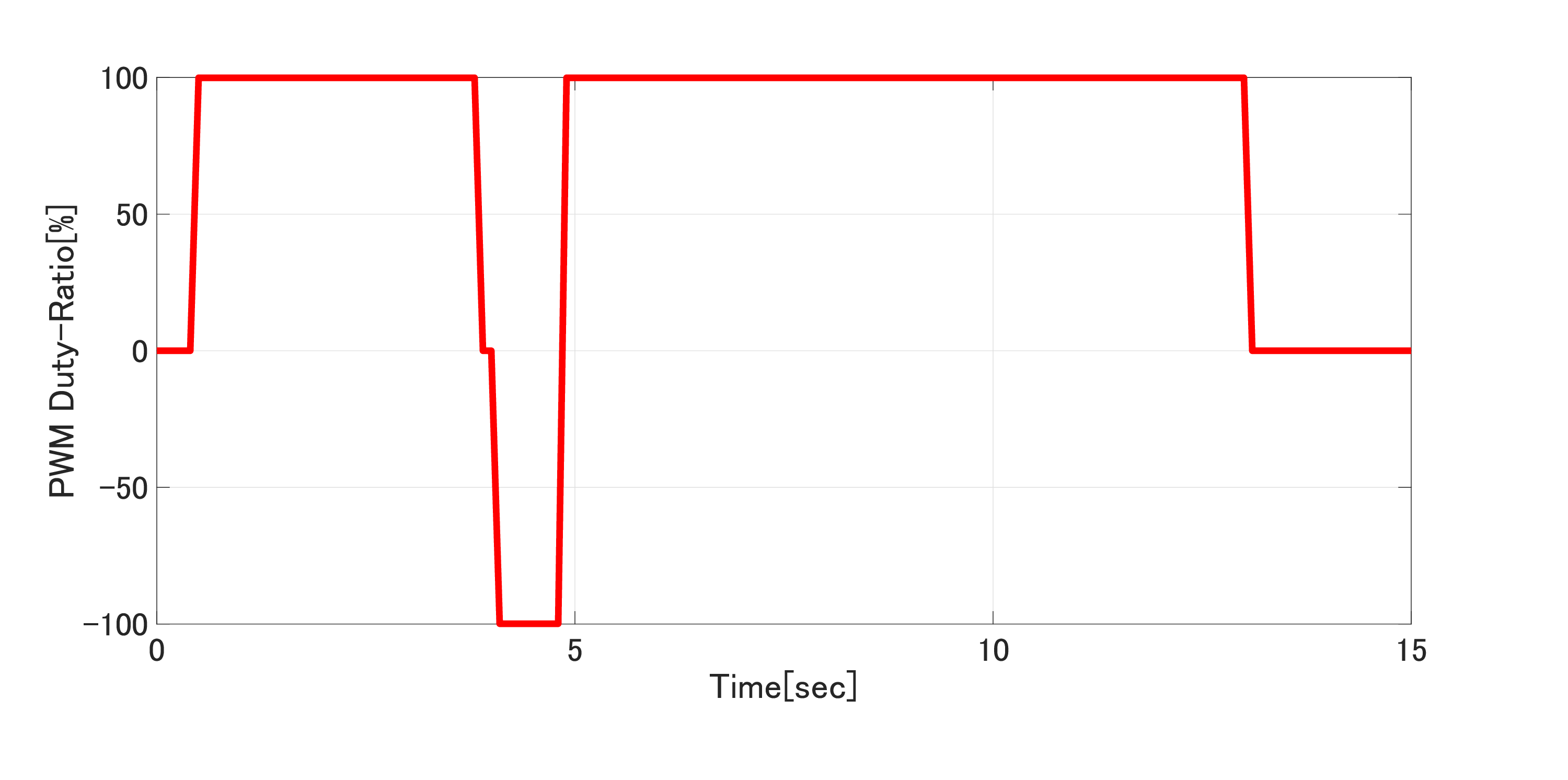}
    \vspace{-25pt}
    \subcaption{Duty ratio of the PWM input}
  \end{minipage}
  \hspace{2mm}
  \begin{minipage}{\columnwidth}
    \centering
    \includegraphics[width=\linewidth]{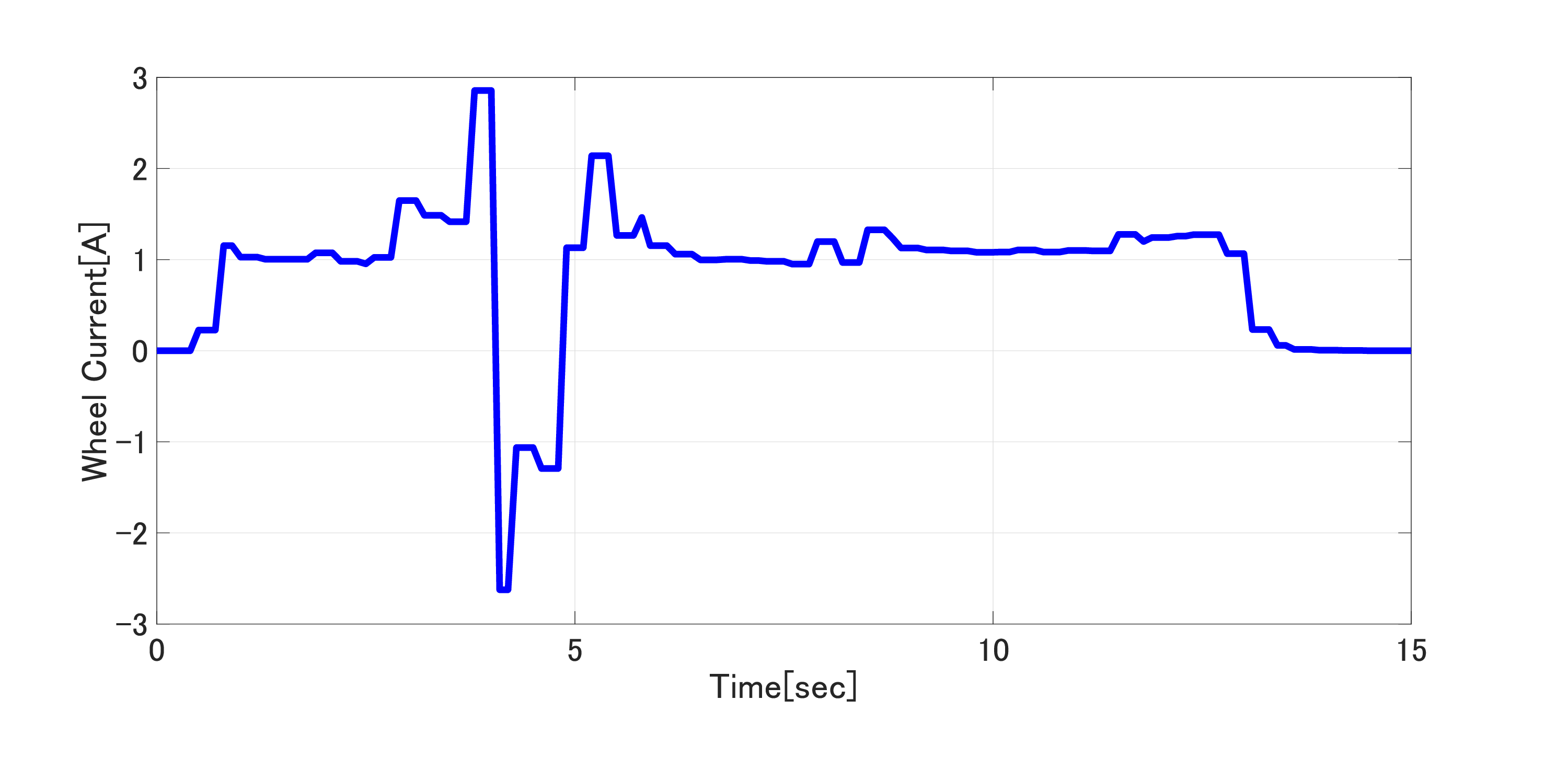}
    \vspace{-25pt}
    \subcaption{Drive wheel motor current}
  \end{minipage}
  \begin{minipage}{\columnwidth}
     \centering
    \includegraphics[width=\linewidth]{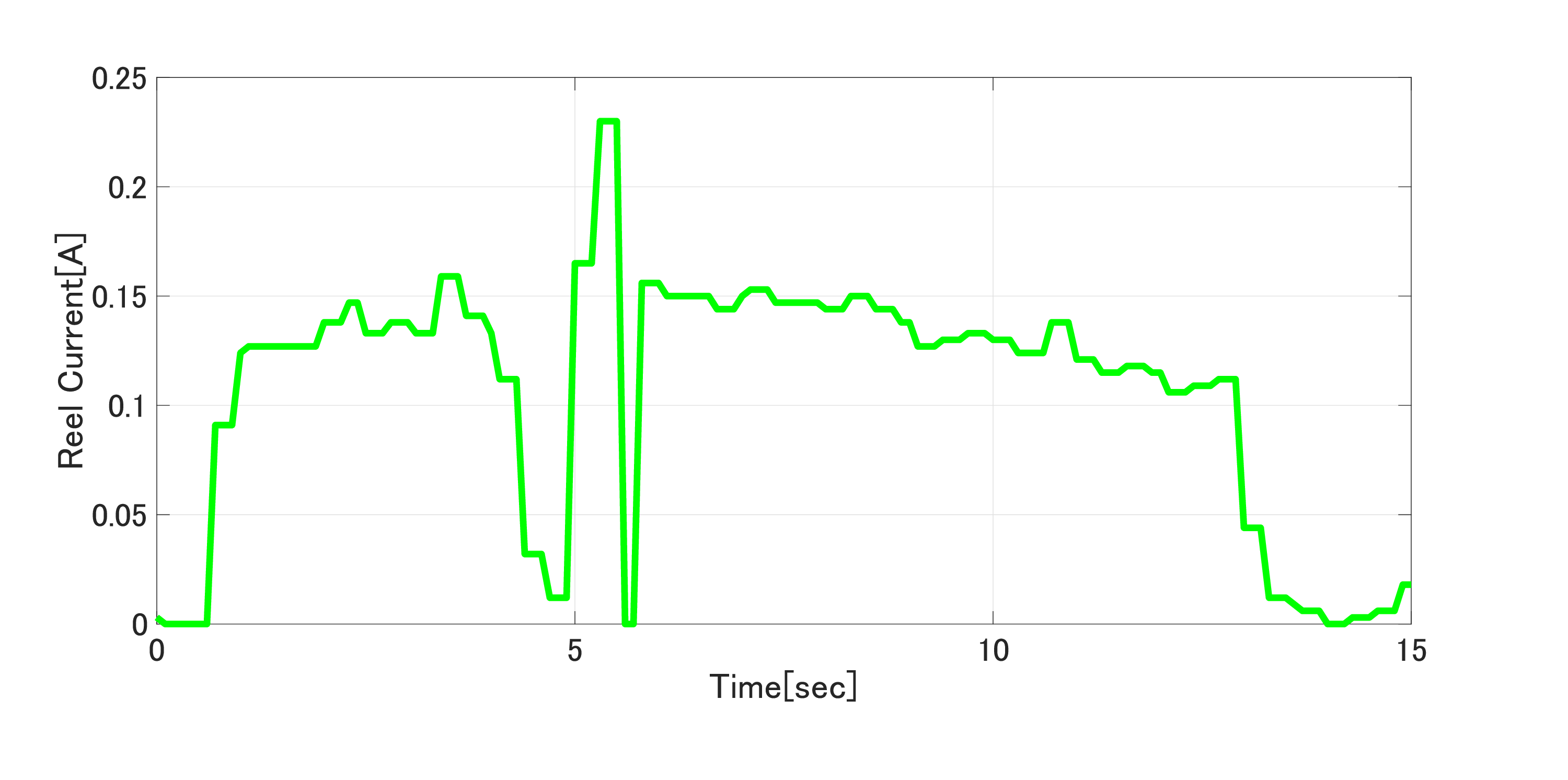}
    \vspace{-25pt}
    \subcaption{Motor current of the winding-up reel}
  \end{minipage}
  \begin{minipage}{\columnwidth}
     \centering
    \includegraphics[width=\linewidth]{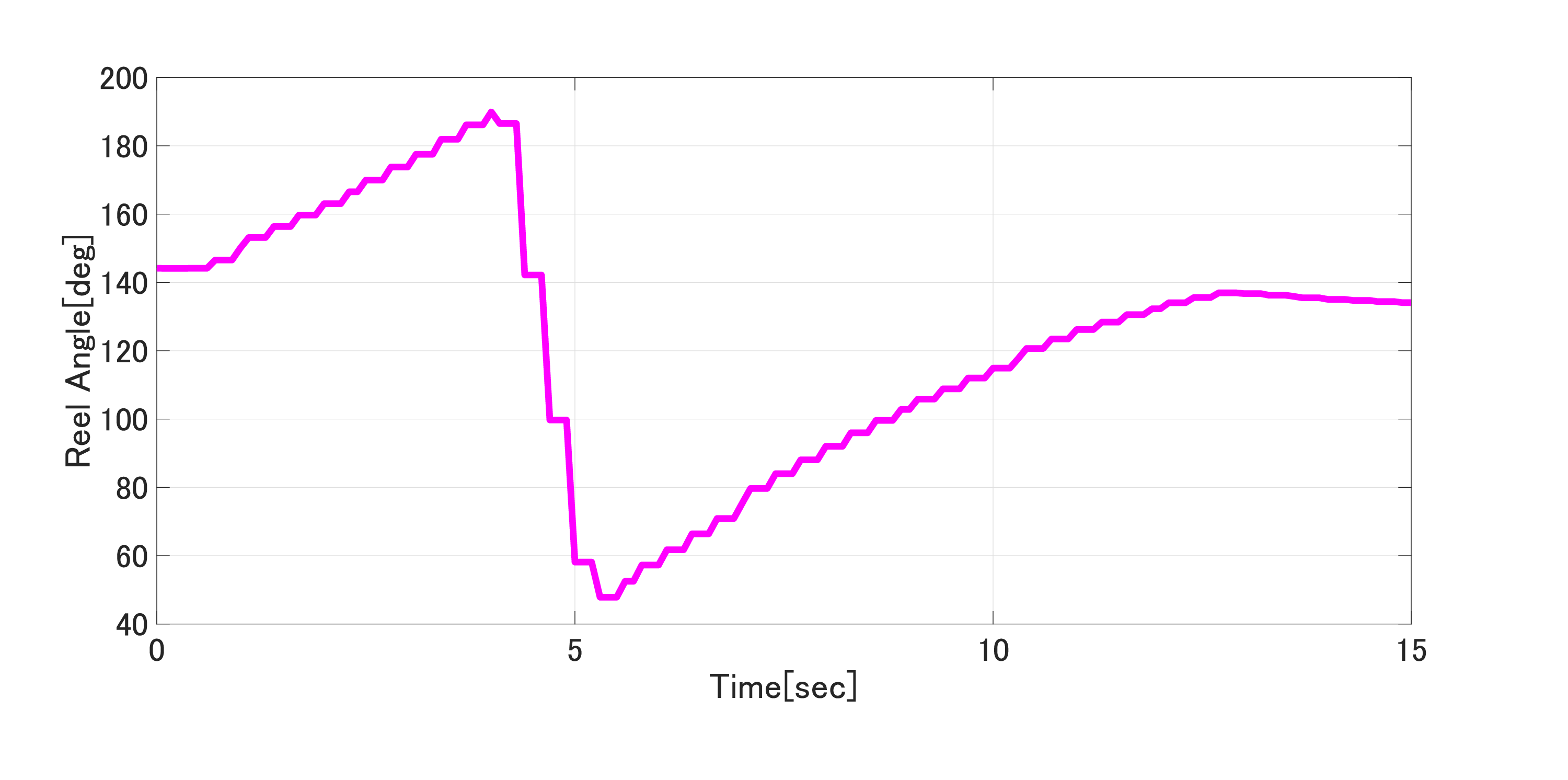}
    \vspace{-25pt}
    \subcaption{Angle of the winding-up reel}
  \end{minipage}
  \caption{Log data in motion with the proposed algorithm for the reel}
  \label{fig:3PropulsionUnit_Exp3_data}
\end{figure}

%段差走破制御プログラムによる自律走行試験
It was confirmed that Xbot-2 with three propulsion units successfully passed through the pipe joint section by applying the aforementioned sequence.
The transition of the actual motion with three propulsion
units and the proposed algorithm of the wire-driven mechanism is shown in Fig. \ref{fig:3PropulsionUnit_Exp3_picture}, and its log data is also plotted in Fig. \ref{fig:3PropulsionUnit_Exp3_data}.
After reaching the pipe joint section, the current rose to approximately 2.9 A at the timing of 3.8 sec.
Since the wheels are driven constantly, the propulsion unit moved backward between 3.9 and 4.4 sec. Furthermore, both the second and third propulsion units smoothly passed through the joint step. Regarding the reel angle, it reversed by 150 deg around 4.0 sec by loosing the wire tension. Travel in the pipe joint step was successfully achieved without getting stuck.
%走行実験を行った結果，配管継手部の走破に成功した．走行の様子をFig.に示し，走行データをFig.に示す．配管継手部に到達後，3.8sec時点で2.9A程度まで電流値が上昇した．走破用制御により，3.9～4.4secで推進ユニットが後退している．また，2基目の推進ユニット，3基目の推進ユニットともに継手段差をスムーズに走破した．リール角度について，4.0sec付近で150°逆転しており，ワイヤが弛緩した．スタックすることなく，配管継手段差の走破に成功した．

%%%%%%%%%%%%%%%%%%%%%%%%%%%%%%%%%%%%%%%%%%%%%%%%%%%%%%%%%%%%%%%%%%%%
\section{CONCLUSION}
In this study, a 3 in sewer pipe inspection robot with an articulated differential mechanism named Xbot-2, which is upgraded from Xbot-1. It has improved traction force from Xbot-1.
It was confirmed that decreasing wire tension is effective for the travel in the pipe joint sections.
Autonomous overcoming for the joint steps was also implemented based on the drive wheel motor current.
The experiment using Xbot-2 with three propulsive units was conducted, and the robot passed through the pipe joint step smoothly. 
The threshold determined experimentally is used to verify the principles of the proposed method, however, a more general step detection method is required for practical application.
In the future, we plan to conduct long-distance inspection in actual piping and verify emergency evacuation performance. 
%本研究では，Xbot-1を改良した連結差動機構を有するXbot-2を開発した．Xbot-1から牽引力が改善していることが確認され，配管継手部走破のためにワイヤの弛緩動作が有効であることが確認された．また，自律走行のための段差走破用制御を実装した．実験を行ったところ，スムーズに通過することが出来た．駆動輪モータ電流値を比較したところ，低負荷で通過できていることが確認された．

%追記：原理検証を目的としたため実験的に閾値を定めたが，実用化に当たり，より一般的な段差接触の検出手法を検討する必要がある．今後，実際の配管における長距離検査や，緊急脱出時の回収性能の検証を実施予定である．

%\section*{APPENDIX}

% \section*{ACKNOWLEDGMENT}
% This study received support from JST University-launched Startup Creation Fund Project KSAC-GAP FUND (Second period), sincerely appriciate their support.

\end{document}